\documentclass[a4paper]{article}

\usepackage{INTERSPEECH2018}
\usepackage{multirow}
\usepackage{multicol}
\usepackage{changepage}
\usepackage{hhline}
 \usepackage{makecell}
 \usepackage[flushleft]{threeparttable}
 \newcolumntype{P}[1]{>{\centering\arraybackslash}p{#1}}
 \usepackage{cite}

\title{Contextual Language Model Adaptation for Conversational Agents}
\name{Author Name$^1$, Co-author Name$^2$}
\name{Anirudh Raju$^1$\textsuperscript{*}, Behnam Hedayatnia$^1$\textsuperscript{*}, Linda Liu$^2$, Ankur Gandhe$^1$, Chandra Khatri$^1$, \\ Angeliki Metallinou$^1$, Anu Venkatesh$^1$, Ariya Rastrow$^1$}

\address{
$^1$Amazon Alexa Machine Learning \\
$^2$ University of Rochester \\
\small\textsuperscript{*}Both authors contributed equally to this work
 }
 
\email{\{ranirudh,behnam,aggandhe,ckhatri,ametalli,anuvenk,arastrow\}@amazon.com}

\begin{document}

\maketitle
\begin{abstract}
Statistical language models (LM) play a key role in Automatic Speech Recognition (ASR) systems used by conversational agents. These ASR systems should provide a high accuracy under a variety of speaking styles, domains, vocabulary and argots. In this paper, we present a DNN-based method to adapt the LM to each user-agent interaction based on generalized contextual information, by predicting an optimal, context-dependent set of LM interpolation weights.  
We show that this framework for contextual adaptation provides accuracy improvements under different possible mixture LM partitions that are relevant for both (1) Goal-oriented conversational agents where it's natural to partition the data by the requested \textit{application} and for (2) Non-goal oriented conversational agents where the data can be partitioned using \textit{topic labels} that come from predictions of a topic classifier. 
We obtain a relative WER improvement of 3\% with a 1-pass decoding strategy and 6\% in a 2-pass decoding framework, over an unadapted model. We also show up to a 15\% relative improvement in recognizing named entities which is of significant value for conversational ASR systems.
\end{abstract}

\vspace{-2mm}
\section{Introduction}
Automatic Speech Recognition (ASR) systems are a key component in building conversational agents. 
The most common approach to building language models (LMs) for ASR systems is to learn n-gram models on large text corpora. These models are trained to predict the conditional word probabilities given the context of the previous $n - 1$ words. Hence, they do not model longer-range dependencies that may vary between different subsets of the training data such as speaking style, vocabulary and topics of conversation \cite{kneser1993dynamic}. 
In this paper, we explore the use of contextual information for adapting the LM to each user-agent interaction. Broadly speaking, any additional information that would help predict the user's speech in the current interaction can be considered as contextual information, including the history of user-agent interactions, meta-data information such as dialog state and time of day, user personalized information like music preferences.

A common approach for LM adaptation is to represent the LM as a mixture of multiple component LMs, where the interpolation weights can be adapted to a single global target domain or dynamically adapted based on contextual information. In order to partition the training data and build the component LMs, a dominant approach has been to use supervised labels in the training data. In \cite{wessel1999,xu2000language,visweswariah2001language}, the \textit{dialog state} was used to build multiple component LMs and in \cite{kneser1993dynamic, suzuki2006}, supervised topic labels were used to build topic-specific component LMs. Other approaches which do not use supervised labels to partition the training data, rely on the availability of in-domain data to adapt LMs by selecting training data that minimizes cross-entropy \cite{axelrod2011domain}, use a combination of mixture-based and MAP-based models \cite{chen2001language} or use constrained KL divergence between unigram distributions \cite{kneser1997language}. Alternatively, clustered models are created either based on latent semantic analysis \cite{bellegarda2000exploiting}, k-means clustering \cite{clarkson1997language} or through topic clusters computed by Latent Dirichlet Allocation (LDA) \cite{tam2005dynamic}. To estimate the context weights adaptively, past approaches used the expectation-maximization algorithm to learn the interpolation weights for a target sentence\cite{clarkson1997language, thadani2012fly}. More recently, using topic vectors directly in an RNN-LM was introduced in \cite{mikolov2012language}. 
 
Our main contribution is to present a framework for adapting n-gram based LMs under some generalized contextual input. In this paper, we use a   Deep Neural Network (DNN) to estimate the interpolation weights for a mixture of n-gram LMs, but the approach can be extended to other adaptation techniques such as n-gram boosting~\cite{aleksic2015bringing}. Unlike the approach presented in~\cite{mikolov2012language}, an n-gram based adaptation can be easily integrated with an online system.
For goal-oriented conversational interactions we partition our component LMs based on the \textit{application label}, while for non-goal oriented, free-form dialogs we build component LMs based on an estimated \textit{topic label}. For both use cases, our proposed framework outperforms an un-adapted framework, and leads to up to 6\% relative WER reduction. Furthermore, our proposed adaptation leads to significant reduction, up to 15\% relative WER, in recognizing topic-related entities of interest, such as person and location names, that appear in conversational interactions.

The paper is organized as follows. Section \ref{section:interpolated_lm} describes the interpolated LM and our strategies for partitioning the data into component LMs, while Section \ref{section:on-the-fly-fst} describes the on-the-fly LM interpolation. Sections \ref{section:experimental-setup} and \ref{section:results-and-discussion} describe the experimental setup, results and discussion. Finally, we conclude in Section \ref{section:conclusions}.

\vspace{-0.2cm}
\section{Interpolated Language Model}
\label{section:interpolated_lm}

The interpolated language model is represented as a mixture of $C$ component LMs, where the probability of a word sequence is calculated by using an interpolation weight $\lambda_{k}$ corresponding to component $LM_{k}$ in the mixture:

\vspace{-0.2cm}
\begin{equation}
\label{eqn:interpolated_lm}
P_{mix}(w_{i}|h_{i})=\sum_{k=1}^{C}\lambda_{k}P_{k}(w_{i}|h_{i})
\end{equation}
such that $\sum_{k=1}\lambda_{k} = 1$

\vspace{-2mm}
\subsection{Building component LMs}
\vspace{-1mm}
\label{section:component-lm}

Building the component LMs requires splitting the LM training data into $C$ components. The training data may be large transcribed corpora of user interactions or other external text corpora. Data should be partitioned along a meaningful dimension e.g., each component representing a different speaking style, vocabulary, discourse topic or some other nuance of language. In this work, we focus on ASR systems for two broad types of conversational agents, for which we use different data partitioning strategies.

\textbf{Goal-oriented conversational agents:} These agents are designed to support a few specific applications such as asking for weather information, playing music etc. Most interactions with personal assistants fall into this category. These interactions are typically short and comprise of very few dialog turns with the objective of providing the user with the requested information. 
Our goal-oriented interaction data is manually labeled in a straightforward manner based on the requested \textit{application}, for example the sentence `Play a song by the Beatles' would be labeled as `Music' application. This enables splitting the training data based on the \textit{application label}, and training one component LM for each application. We expect that user interactions with a music application would have different n-gram statistics compared to interactions with a shopping application.

\textbf{Chatbots} Chatbots are non goal-oriented conversational agents where the objective is to engage the user in an interesting and coherent conversation, as opposed to completing a specific task. Examples of older chatbots include ELIZA \cite{weizenbaum1966eliza}, while recent systems include the conversational agents that were deployed as part of the Alexa Prize competition \cite{alexa_prize}. Non goal-oriented human-chatbot dialogs typically contain a variety of conversation topics, therefore we choose the \textit{topic label} as a natural way to partition the data for building the component LMs. Our data is not manually annotated with topic labels, so we use an off-the-shelf topic classifier designed for estimating conversational topics \cite{guo2018topic}. In future, this can be extended by unsupervised clustering of the training data.

\vspace{-2mm}
\subsection{Building component LMs on the fly}
\vspace{-1mm}
\label{section:on-the-fly-fst}
In weighted finite state transducers (WFST) based ASR systems
\cite{mohri2012wfst}, n-gram models are represented as WFSTs where each state represents the n-gram history $h$ and the weight on an out-going arc is either the word probability $p(w_{i} | h)$ or the back-off weight $\alpha{h}$. For each input utterance, an interpolated n-gram LM needs to be built based on interpolation weights. The proposed models for estimating the interpolation weights based on context are described in Section \ref{section:estimate_weights}. After the weights are computed dynamically for each input utterance, we use an efficient on-the-fly interpolated WFST strategy \cite{ballinger2010} where the n-gram probabilities from each component LM are kept separate and interpolated only at run-time.

\vspace{-0.2cm}
\section{Estimation of interpolation weights}
\vspace{-0.2cm}
\label{section:estimate_weights}
Given the interpolated LM representation as described in Equation \ref{eqn:interpolated_lm}, we need to estimate the interpolation weights $\lambda_{k}$ during inference for each utterance based on the contextual information available. We propose to use a DNN model that estimates the $\lambda_{k}$'s as the output of a softmax function, given any generic contextual feature input. The model can be trained in either a supervised or semi-supervised manner, as described below.

\vspace{0.1cm}
\textbf{Minimize LM perplexity:} The contextual adaptation model can be trained to estimate the interpolation weights $\lambda_{ik}$ such that it maximizes the log-likelihood of each utterance in the training data under the interpolated model. This is equivalent to minimizing the LM perplexity of the training data. For an utterance $\vec{u_{i}} = (w_1, w_2....w_N)$ of $N$ words, the perplexity-based loss function (PPL) is:

\vspace{-0.4cm}
\begin{align}
\label{eqn:min_perplexity}
Loss  &= - log  p(w_{1}, w_{2}...w_{N})\\
         &= - \sum_{N} log  p(w_{j} | w_{j-1},...w_{j-n-1}) \\
         &= - \sum_{N} log  \sum_{K} \lambda_{k} p_{k}(w_{j}|w_{j-1},...w_{j-n-1})
\end{align}
where $p_k$ is the probability of the n-gram from $LM_{k}$. We compute the derivative of our loss function w.r.t. to $\lambda_{k}$ and back-propagate the error to estimate the DNN parameters. To efficiently compute the loss, we pre-calculate $p_{k}(w_{j} |w_{j-1},...w_{j-n-1})$ for all examples as it stays constant throughout the optimization.

The training data can come from text corpus or from user-agent interactions. The user-agent interaction data can either be ASR 1-best recognitions (resulting in a semi-supervised training) or manually transcribed data (supervised training). 
\newline

\vspace{-0.2cm}
\textbf{Minimize cross-entropy loss for component LM labels:} 
The contextual adaptation model is trained to directly predict the intended component LM label for each utterance in the training data, using a cross-entropy (xent) loss function. The target labels come from manual annotations of the requested \textit{application} for goal-oriented conversational agents, and from topic label estimates obtained from a topic classifier \cite{guo2018topic} for chatbots.

\vspace{-0.2cm}
\section{Experimental Setup}
\label{section:experimental-setup}
We build two separate ASR/LM systems, for goal-oriented and non goal-oriented conversations respectively. Section \ref{section:datasets} describes the datasets used for training the LMs, while Section \ref{section:asr_systems} provides details on LM building and the context adaptation model for each of the two cases.

\vspace{-0.1cm}
\subsection{Datasets}
\label{section:datasets}
We have two datasets of user interactions with a conversational agent, specifically \textit{goal-oriented} and \textit{non goal-oriented} interactions, which are used for LM training and for evaluation. For our experiments, both datasets are split into the following partitions - 80\% train, 10\% dev, and 10\% test. Additionally, we also have access to large external text corpora that are used only for LM training, and not as test sets.

\textbf{Goal-oriented interaction data:} This dataset consists of millions of utterances collected in far-field conditions from real user interactions with Alexa, a goal-oriented conversational agent. Each utterance corresponds to a single turn of user-agent interaction that has been annotated with the \textit{application} that was requested and the corresponding \textit{text transcription}. Application labels include Music, Shopping, Weather and others.

\textbf{Non goal-oriented, chatbot interaction data:} This dataset consists of hundreds of thousands of far-field speech-based user-agent interactions with a chatbot. The goal of the chatbot is to engage with the user in a conversation. The data is bucketed into \textit{conversations}, where a single conversation is initiated and terminated by the user, and consists of multiple turns of user-agent interactions.

\textbf{External text datasets} We use a variety of external text corpora, from multiple sources such as news, voice-mail, web crawled corpora, etc. The total data size is of the order of billions words and it is entirely in the train partition.

\vspace{-0.2cm}
\subsection{ASR Systems}
\label{section:asr_systems}
For all our experiments we use an experimental ASR system that does not reflect the performance of the production Alexa system. We build two ASR systems - one for goal-oriented agents and another for chatbots, each containing component LMs trained on different data, as described in Section \ref{section:component_lms_setup}. The structure of the LMs in both systems is a mixture of Katz smoothed \cite{Katz1987} 4-gram language models which are interpolated on-the-fly in a WFST decoding framework, as described in Section \ref{section:on-the-fly-fst}. We also build baseline unadapted ASR systems for each of these use cases, by estimating static interpolation weights to minimize perplexity of the corresponding dev set.

\vspace{-0.1cm}
\subsection{Component LMs}
\label{section:component_lms_setup}
\begin{table}
 \caption{Description of LM setup in each ASR system. Data sources are described in Section \ref{section:datasets}}
 \label{table:component-lms}
\begin{tabular}{l l l}
\hline
\multirow{2}{*}{\textbf{LM details}} & \multicolumn{2}{c}{\textbf{ASR System}}\tabularnewline
 & \textbf{Goal-oriented} & \textbf{Chatbot} \\
\hline 
Component LMs  & Application & Topic \\
\hline 
No. Comp. LMs & 13 & 26 \\
\hline 
Training data & goal-oriented & chatbot and \\
 & data & external data \\
\hline
\vspace{-0.9cm}
\end{tabular}
\end{table}

In Table \ref{table:component-lms}, for each ASR system, we describe the training data used to build the LMs, the number of component LMs, and what they represent (application vs topic label). LM training is described in Section \ref{section:component-lm}. The goal-oriented ASR system uses the annotated application label to partition the LM training data, while the chatbot LM uses the topic labels estimated from a topic classifier \cite{guo2018topic}. Since the topic labels are obtained from a classifier as opposed to manual annotations, this scales easily and allows us to use large external text corpora in addition to user-agent interaction data, for training the chatbot LM. We use the following strategy to mix data from multiple external data sources for the chatbot LM -  for each of the component topic based LMs, we build data source specific LMs which are statically interpolated to minimize perplexity on the corresponding dev set, i.e., dev partitions of the non goal-oriented chatbot conversations (Section \ref{section:datasets})

\vspace{-0.2cm}
\subsection{Contextual adaptation model}
\begin{figure}[t]
 \vspace{-0.5cm}
  \centering
  \includegraphics[width=0.9\linewidth]{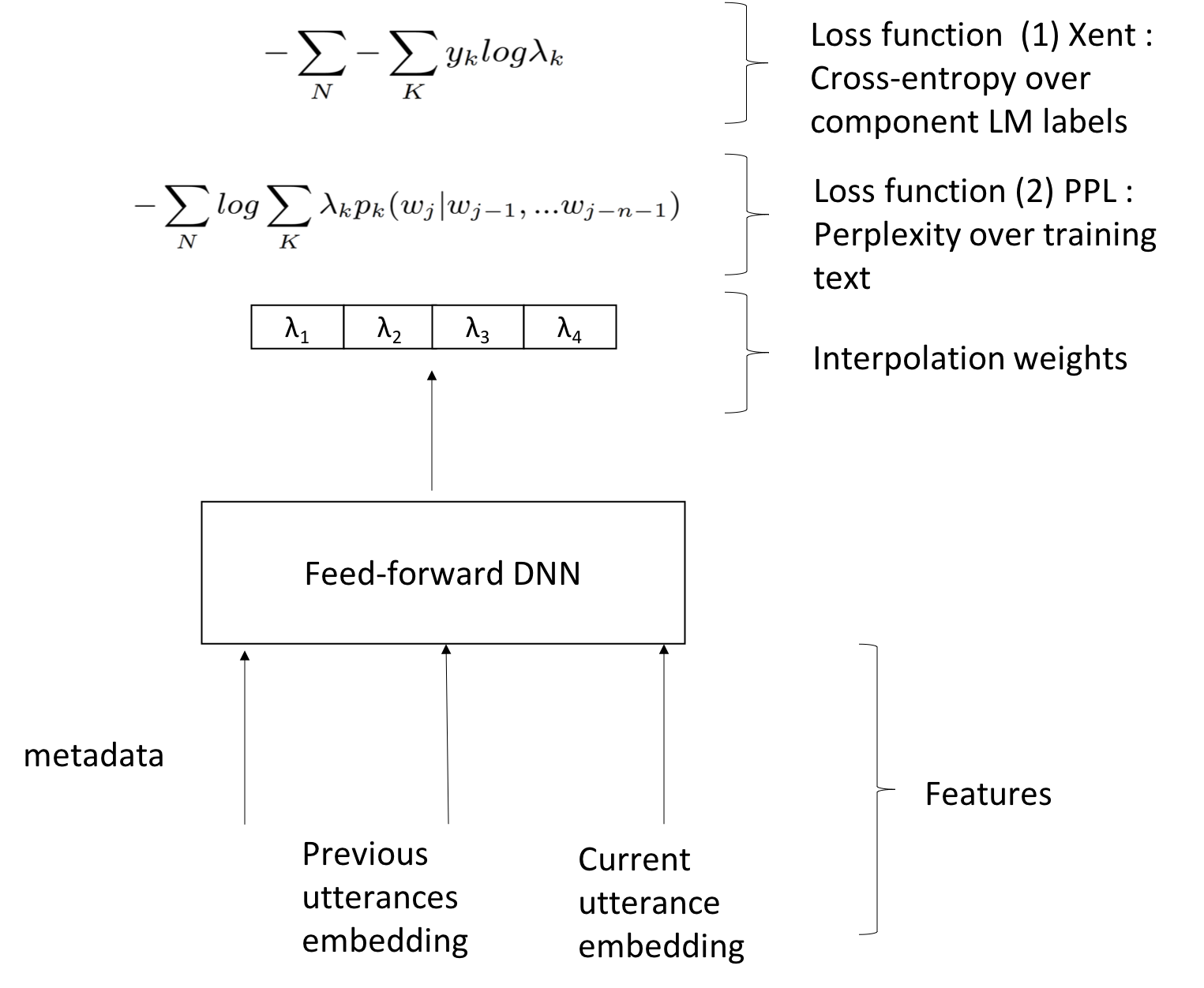}
  \caption{Deep Neural Network (DNN) based contextual adaptation model to predict interpolation weights}
  \label{fig:model}
\vspace{-0.7cm}
\end{figure}

The contextual adaptation model for each ASR system is trained from their respective user-agent conversational datasets (goal-oriented and non goal-oriented datasets of Section \ref{section:datasets}). We use a few hundred thousand utterances for training the contextual models. As described in Section \ref{section:estimate_weights}, we train the models to either minimize cross-entropy (xent) of the component LM label distribution or perplexity (PPL) of the training data text. The DNN model shown in Figure \ref{fig:model} is a two layer network of 200 hidden units which is trained using the Adam optimizer, clipping gradients and early stopping. The $\lambda_{k}$ parameters are estimated from the final softmax layer. 

\vspace{-0.1cm}
\subsubsection{Contextual Features}
The DNN-based contextual adaptation model allows for generic contextual features. The ones that we experiment with are obtained either from (1) a context window of past user-agent interactions or (2) the metadata information of the current interaction, i.e. time of day. Note that, these features are available prior to each user-agent interaction. The estimates from the DNN are used to build an adapted LM for each utterance, using which we run a 1-pass decoding with a real-time ASR system.

In addition, we report 2-pass decoding experiments where we use another feature  - (3) 1-best hypothesis from the current interaction. The adapted LM estimated in this case is used to re-decode the utterance. This is suitable for a non-real time system, and provides an upper bound on the possible WER reduction, since the adapted LM is used as part of beam search decoding. Note that, in order to deploy this into a real-time system, we can run the first-pass decoding using the adapted LM from past utterance features and subsequently rescore the lattice using the more powerful adapted LM which includes current utterance features. This would result in a WER reduction in-between the 1-pass and 2-pass results reported here.

The context window that we used for the past utterance features depends on the type of conversation. For goal-oriented conversational agents, we use all user-agent interactions within the past $T$ seconds of the current interaction. For non goal-oriented conversational agents, we use all user-agent interactions within the current conversation.

We have three top level features: past interactions based features (shorthand : \textit{prev}), metadata based features (shorthand : \textit{meta}) and current interaction based features (shorthand : \textit{cur}). \textit{prev} features include averaged pre-trained word embeddings~\cite{pennington2014glove} of all previous \textit{user} turns within the context window and averaged word embeddings of all previous \textit{agent} responses within the context window. \textit{meta} features include the day of week and the time of day (morning, afternoon, evening). Lastly, \textit{cur} features include the averaged word embeddings from 1-best ASR recognition of the current utterance.

\vspace{-0.2cm}
\section{Results and Discussion}
\label{section:results-and-discussion}
 \vspace{-3mm}
\begin{table}[th]
\footnotesize
  \caption{Perplexity (PPL) and relative WER reduction (WERR) for the goal-oriented ASR system on a goal-oriented test set, for different objective functions and features for the DNN context model}
  	\vspace{-3mm}
  \label{table:goal-results}
  \centering
   \begin{adjustwidth}{0cm}{}
  \begin{tabular}{l l l l l} 
    \toprule
\textbf{Model} & \textbf{Feats} & \textbf{PPL} & \textbf{WERR(\%)} & \textbf{Entity}  \\ 
 & & & & \textbf{WERR(\%)}  \\ 
    \hline
	  \multicolumn{4}{c}{\textit{decoder : 1-pass}} \\
	 \hline
    No Adapt & -  & 30.66 & - & -\\

  DNN(Xent) & prev, meta  & 29.43 & +0.33\%  & +1.17\% \\

 DNN(PPL) & prev, meta & 26.10  & -1.25\% & +0.21\%  \\
      	\hline
	  \multicolumn{4}{c}{\textit{decoder : 2-pass}} \\
	 \hline
  DNN(Xent) & prev, meta, cur & 20.63 & -3.49\% & -3.04\%   \\

  DNN(PPL) & prev, meta, cur & 20.30 & -3.24\% & -2.21\%   \\
   \bottomrule
  \end{tabular}
  \end{adjustwidth}
\end{table}
 \vspace{-0.4cm}

\vspace{-2mm}
\subsection{Overall results using both 1-pass and 2-pass decoding}
Table \ref{table:goal-results} presents the perplexity and relative Word Error Rate (WER) reductions for the goal-oriented ASR system on a goal-oriented test set. We observe a relative WER reduction of 1.25\% relative in a single-pass decoding framework and 3.5\% relative in a 2-pass decoding framework by using the context adaptive LM. Similarly, in Table \ref{table:chatbot-results}, we present the results of the chatbot ASR system on a chatbot test set, where we see similar trends. Specifically, we observe a 2.7\% rel. WER reduction in single-pass and a 6\% rel. WER reduction in the 2-pass framework. The improved results for the 1-pass system are promising because they show that we can predict future behavior based on usage history and other metadata. As expected, the WER results are better in a 2-pass framework because knowledge of the current utterance helps us build a better adaptive LM. The current utterance features are the strongest signal to help improve WER and hence result in maximum improvements. However, this comes at the cost of higher latency of the 2-pass system compared to 1-pass.

\vspace{-2mm}
\subsection{Impact of different loss functions}
From Tables \ref{table:goal-results} and \ref{table:chatbot-results} for both goal-oriented and chatbot ASR systems, we observe that when we train the context adaptation model using the perplexity (PPL) objective instead of cross-entropy (xent), we achieve adapted LMs with better PPL and WER on the test set. Moreover, training with the PPL objective function has the advantage of not requiring explicit labels of the optimal component LM for each utterance, and can be extended to scenarios where we may wish to train with semi-supervised data as described in Section \ref{section:estimate_weights}. 

\begin{table}[th]
\footnotesize
  \caption{Perplexity (PPL) and relative WER reduction (WERR) for the chatbot ASR system on a chatbot test set, for different objective functions and features for the DNN context model}
  	\vspace{-3mm} 
\label{table:chatbot-results}
\centering
\begin{adjustwidth}{-0.1cm}{}
  \begin{tabular}{l l l l l} 
    \toprule         
\textbf{Model} & \textbf{Feats} & \textbf{PPL} & \textbf{WERR(\%)} & \textbf{Entity}  \\ 
 &  &  & & \textbf{WERR(\%)}  \\ 
	\hline
	  \multicolumn{4}{c}{\textit{decoder : 1-pass}} \\
	 \hline
  No Adapt  &  -   &60.77 & - & - \\

     DNN (Xent) &  prev, meta  & 59.81 & -1.73\% &  -8.19\% \\

  DNN (PPL) &  prev, meta  & 58.14 & -1.61\% &  -2.98\%  \\
 
  DNN (PPL) &  prev-d, meta  & 55.66 & -2.76\% & -10.92\% \\
  	\hline
	  \multicolumn{4}{c}{\textit{decoder: 2-pass}} \\
	 \hline
   DNN (PPL)  & prev, cur, meta  & 42.03 & -5.58\% & -15.15\% \\

  DNN (PPL)  & prev-d, cur, meta  & 42.83 & -5.92\% & -14.67\% \\

  DNN(PPL)  & cur, meta   & 42.72 & -5.98\% & -15.32\% \\

  Topic model  & cur  & 45.08 & -5.52\% & -13.14\% \\
  
  \bottomrule
\end{tabular}
\end{adjustwidth}
\end{table}
 \vspace{-0.2cm}

\vspace{-2mm}
\subsection{Impact of decaying past context}
An interesting observation from our data is that users tend to interact with a chatbot in multiple turns, while goal-based interactions are significantly shorter. The typical context window for the chatbot system contains several more user-agent turns than for the goal-based system. Based on this insight, we further improved our contextual model by slightly modifying the contextual features used in the lengthy chatbot interactions. Specifically, we performed an exponentially decaying weighted average over the past word-embeddings, to give higher weight to recent utterances that are closer to the current utterance (called \textit{prev-d} in Table \ref{table:chatbot-results}). Using this exponential decay improves over using the standard \textit{prev} features in both PPL and WER. In Table \ref{table:chatbot-results}, using decaying context features (\textit{prev-d}) leads to an overall WER reduction of 2.76\% using the context adapted LM in a single-pass decoding framework and 5.92\% in a 2-pass decoding framework.

\vspace{-2mm}
\subsection{Comparison with topic classifier based adaptation}
We compare results when using the weights from the context adaptation model vs weights directly estimated from a topic classifier in a 2-pass decoding framework, see topic model vs DNN (PPL) in Table \ref{table:chatbot-results}. For the topic model, the topic LM interpolation weights are the final output topic probabilities estimated by the topic classifier, which is the same classifier we used to partition the data into component LMs (\cite{guo2018topic}, Section \ref{section:component_lms_setup}). From Table \ref{table:chatbot-results}, we observe that using this strategy, leads to competitive results as compared to the context adaptation model that optimizes for PPL, i.e., leading to a 5.5\% relative WER reduction compared to the chatbot baseline. 

\vspace{-2mm}
\subsection{Impact on Named Entity accuracy}
While WER is a standard metric to measure the performance of ASR systems, it does not capture the relative importance of different words in an utterance. For conversational systems, named entities such as people names, locations etc, are arguably more important compared to other words due to their impact on downstream tasks such as Natural Language Understanding (NLU). These entities tend to be topic specific, e.g., conversations about music contain entities such as artist and song names. Here, we analyze the accuracy of our proposed contextual ASR system on named entities. To measure entity error rate, we tag each word in our test data using an in-house Named Entity Recognition (NER) tagger. The entity error rate is defined as $(num\_substitutions + num\_deletions)/num\_reference-words$. Note that we do not include insertions due to difficulty in attributing whether an insertion error was caused by the entity or the other surrounding words.

In the goal-oriented system, we bias towards the next application that the user is likely to request such as Music or Weather. Hence, in Table \ref{table:goal-results}, we see similar improvements in the carrier phrase recognition as well as the entities i.e. the overall WER reduction is similar to the entity WER reduction. In contrast, the non-goal oriented data is more challenging due to it's free-form conversational nature, and contains topic specific named entities. We see a larger reduction of 15.32\% entity error rate here in Table \ref{table:chatbot-results} with the best adapted LM. For example, difficult entities like "czechoslovakia" and "abigail" are correctly recognized by the adapted LM. This is a promising result for conversational ASR systems, where successful recognition of named entities is critical for completing the user request or engaging the user in meaningful dialog about topics of interest.

\vspace{-0.2cm}
\section{Conclusions and Future Work}
\label{section:conclusions}
We described methods for improving the performance of a mixture of n-gram LMs with on-the-fly interpolation, using a contextual adaptation framework. We presented a DNN-based method to predict an optimal set of interpolation weights for each interaction, in an online fashion, from generalized contextual information. This model was evaluated for both goal-oriented conversational agents where we partitioned the data by the requested application and on non-goal oriented conversational agents where we partitioned the data using topic labels that come from predictions of a topic classifier. We achieved a relative WER reduction up to 3\% in a 1-pass decoding strategy using past context only, and up to 6\% relative in a 2-pass decoding strategy, over a statically interpolated baseline LM. Our method benefits named entities the most: we reduced entity error rate by up to 15.32\% relative. Future work includes evaluating this contextual LM adaptation strategy using other relevant context, such as topics derived from unsupervised clustering of the conversation, multimodal information that is available to the conversational agent or information about current events.

\bibliographystyle{IEEEtran}

\bibliography{mybib}
 
\end{document}